\documentclass{article}
\usepackage{ijcai26}

\usepackage{times}
\usepackage{soul}
\usepackage{url}
\usepackage[hidelinks]{hyperref}
\usepackage{multirow}
\usepackage[utf8]{inputenc}
\usepackage[small]{caption}
\usepackage{graphicx}
\usepackage{amsmath}
\usepackage{amssymb}
\usepackage{amsthm}
\usepackage{booktabs}
\usepackage{algorithm}
\usepackage{algorithmic}
\usepackage[switch]{lineno}
\usepackage{array}
\usepackage{makecell}

\title{Knowing When Not to Predict: Self Supervised Learning and Abstention for Safer DR Screening}

\author{
Muskaan Chopra$^{1, 4}$
\and
Lorenz Sparrenberg$^{1, 4}$\and
Jan H. Terheyden$^2$\and
Rafet Sifa$^{1, 3, 4}$\\
\affiliations
$^1$Rheinische Friedrich-Wilhelms-Universität Bonn, Bonn, Germany\\
$^2$University Hospital Bonn - Department of Ophthalmology, Germany\\
$^3$Fraunhofer IAIS, Sankt Augustin, Germany\\
$^4$Lamarr Institute for Machine Learning and Artificial Intelligence, Germany\\
}
\begin{document}

\maketitle

\begin{abstract}
Self-supervised learning (SSL) is now a standard way to pretrain medical image models, but performance is still mostly judged by downstream accuracy. For safety-critical screening tasks such as diabetic retinopathy grading, this is not enough: a model must also know when its predictions are unreliable and defer uncertain cases for clinical review. In this work, we examine how the length of SSL pretraining influences calibrated confidence and confidence-based abstention.
We evaluate multiple SSL checkpoints under a fixed fine-tuning protocol and assess calibrated confidence, coverage, selective accuracy, and selective macro-F1. Across datasets and data regimes, SSL pretraining improves selective prediction compared to training from scratch. Unlike prior SSL studies that primarily evaluate downstream accuracy or AUROC, we analyze how SSL pretraining duration influences confidence behavior under calibrated confidence-based abstention. However, once accuracy saturates, selective performance can still change markedly across checkpoints, and longer pretraining does not consistently improve reliability. These results underscore the importance of abstention-aware evaluation and suggest that pretraining length should be treated as an important reliability-related design choice rather than only a computational detail. Code is available at \url{https://github.com/muskaan712/ijcai-knowing-when-not-to-predict}.
\end{abstract}

\section{Introduction}

Diabetic retinopathy (DR) screening is a safety-critical clinical task. In a screening workflow, a model that makes a confident but incorrect prediction can be more harmful than a model that defers uncertain cases to expert review. In real-world settings, screening systems must handle non-gradable images, borderline disease stages, and variability in labeling and grading protocols \cite{gulshan2016jama,krause2018ophthalmology,tsiknakis2021reviewdr}. These factors make reliability, not only average accuracy, essential for clinical deployment.

Self-supervised learning (SSL) has become a common approach for representation learning in medical image analysis, particularly when labeled data are limited \cite{albelwi2022survey,chen2020simclr,he2020moco,grill2020byol,zbontar2021barlow,bardes2022vicreg}. In ophthalmology and fundus-based DR grading, SSL and transfer learning are widely used to improve performance under limited annotation budgets \cite{li2018transferdr,tsiknakis2021reviewdr}. Despite this adoption, evaluation in medical imaging studies still largely centers on downstream accuracy or AUROC \cite{wang2021review}. Such metrics provide an incomplete picture for screening applications, where uncertainty and referral decisions play a central role.

Selective prediction and calibration provide a more deployment-aligned evaluation paradigm. Under selective prediction, the model predicts only for high-confidence cases and abstains on uncertain inputs \cite{geifman2017selective}, while calibration aligns predicted confidence with empirical correctness \cite{guo2017calibration}. These components are often applied post hoc to a trained model, and representation learning choices are rarely examined through the lens of selective behavior, despite their direct influence on confidence distributions.

A frequently overlooked design aspect is SSL pretraining length. In practice, the number of pretraining epochs is often set by compute budgets or marginal gains in downstream accuracy after performance saturates. Whether this choice also affects reliability properties such as calibration and selective prediction remains unclear, even though they are central to screening workflows.

In this work, we study SSL pretraining length as a reliability factor and analyze how the duration of pretraining affects downstream confidence behavior and selective prediction in DR screening. We evaluate checkpoints spanning early, intermediate, and late pretraining stages, fine-tune each under an identical supervised protocol, and assess both standard performance and reliability-aware metrics under confidence-based abstention \cite{geifman2017selective,guo2017calibration}.

We use APTOS as the primary benchmark \cite{tsiknakis2021reviewdr} and additionally evaluate transfer to Messidor and a 7-class fundus dataset using the same pipeline. To isolate representation-level effects on confidence behavior, CAM-based refinement \cite{zhou2015learningdeepfeaturesdiscriminative} is excluded from selective prediction experiments.

Our main finding is that longer SSL pretraining does not monotonically improve reliability. While downstream accuracy often saturates early, selective macro-F1 and confidence behavior can vary substantially across checkpoints, and in some cases selective performance decreases at clinically relevant operating points. These effects are not apparent from accuracy alone and emerge only under selective prediction, which better reflects real screening use.

\paragraph{Contributions.}
We make three contributions. First, we analyze how SSL pretraining duration affects reliability-aware evaluation in diabetic retinopathy screening. Second, we evaluate checkpoints using selective prediction metrics, including coverage, selective accuracy, and selective macro-F1. Third, using fixed downstream splits, calibration, and training settings, we isolate representation-level effects on confidence behavior.

\section{Related Work}

Our work bridges two areas that are often studied separately: self-supervised representation learning for medical imaging, and reliability-aware evaluation via calibration and selective prediction. We summarize relevant work and position our contribution in diabetic retinopathy (DR) screening.

\subsection{Self-Supervised Learning in Medical Imaging}

Self-supervised learning (SSL) learns visual representations from unlabeled data. Contrastive methods such as SimCLR and MoCo use positive pairs from augmented views and negatives from other images \cite{chen2020simclr,he2020moco}, while non-contrastive methods avoid explicit negatives through architectural asymmetry or redundancy reduction, for example BYOL, Barlow Twins, and VICReg \cite{grill2020byol,zbontar2021barlow,bardes2022vicreg}. Surveys review these families and their medical imaging adaptations \cite{albelwi2022survey,wang2021review}.

SSL has been applied across medical imaging modalities, including histopathology, X-ray imaging, and retinal fundus photography \cite{ciga2022histopathologyssl,shen2019adversarialxray,li2021milcovidssl}. In ophthalmology, SSL is widely used for DR grading under limited labels and domain shift \cite{li2018transferdr,tsiknakis2021reviewdr}. However, evaluation still largely focuses on downstream accuracy or AUROC rather than reliability under selective prediction.

\subsection{Calibration and Selective Prediction}

Confidence calibration aims to align predicted probabilities with empirical correctness \cite{guo2017calibration}. Selective prediction allows models to abstain on low-confidence inputs, trading coverage for reduced risk \cite{geifman2017selective}. Learned abstention methods such as SelectiveNet optimize coverage jointly with prediction \cite{geifman2019selectivenet}, but add modeling and optimization complexity. A widely used baseline is confidence-threshold abstention, often applied after calibration, following the classical reject-option rule \cite{chow1970,franc2023}.

In medical imaging, selective prediction is closely related to human-AI collaboration and referral workflows, including DR screening systems that defer uncertain or ungradable cases \cite{defauw2018clinically}. Still, many studies treat abstention as a post hoc decision layer, and the interaction between representation learning choices and resulting confidence distributions is not analyzed in depth \cite{angelopoulos2024,cattelan2023}.

\subsection{Positioning of This Work}

We treat selective prediction performance as an emergent downstream property influenced by representations learned during SSL pretraining. Abstention is not part of pretraining in our setup. We sample checkpoints at different pretraining lengths, fine-tune each checkpoint under an identical downstream protocol, and evaluate confidence-threshold abstention using calibrated confidence and selective metrics. Rather than focusing primarily on algorithmic novelty in SSL or abstention mechanisms, we study how SSL pretraining duration affects calibration and risk-coverage behavior in DR screening, addressing an evaluation gap that is directly relevant to deployment.

\section{Methodology}

We study how the length of self-supervised pretraining influences downstream confidence behavior and selective prediction in diabetic retinopathy screening. To isolate the effect of representation learning, we adopt a controlled protocol in which only the SSL checkpoint varies, while downstream training, calibration, and evaluation settings are kept fixed.

We consider two self-supervised objectives under a common pipeline: SiCoVa (Self-Informed Cross-Correlation for Variance and Invariance), a hybrid non-contrastive objective combining variance, invariance, and cross-correlation regularization, and a contrastive triplet-loss baseline.

\subsection{Pretraining Setup}

Self-supervised pretraining is performed on unlabeled EyePACS fundus images. From each input image $x$, we generate two stochastic views $v_1=t_1(x)$ and $v_2=t_2(x)$ using standard augmentations, including random cropping, horizontal flipping, color jittering, and Gaussian blur. A lightweight Jigsaw perturbation \cite{jigsaw} is applied at the input to encourage patch-level discrimination and fine-grained spatial sensitivity.

Each view is processed by a shared encoder $f_{\theta}$, implemented as either a ResNet50 \cite{resnet50} or a ViT-B/16 \cite{vit}, followed by a three-layer MLP projector $g_{\phi}$. For any view $v$, this produces an embedding
\[
z = g_{\phi}\!\left(f_{\theta}(v)\right).
\]
For a mini-batch of size $N$, we write the paired embeddings as $Z, Z' \in \mathbb{R}^{N \times D}$, where the $i$-th row corresponds to the $i$-th image in the batch, for $i \in \{1,\ldots,N\}$. Concretely, we denote the embeddings of the two views of image $x_i$ by
\[
z_i = g_{\phi}\!\left(f_{\theta}(v_{1,i})\right), \qquad
z'_i = g_{\phi}\!\left(f_{\theta}(v_{2,i})\right).
\]
During SSL training, checkpoints are saved at regular intervals spanning early, intermediate, and late stages of convergence. Since all checkpoints share the same architecture and pretraining configuration, observed downstream differences are primarily associated with pretraining length, although some variability due to stochastic optimization and fine-tuning randomness is expected.

\subsection{Non-Contrastive SSL with SiCoVa}

SiCoVa is a non-contrastive objective that encourages informative, non-collapsing, and decorrelated representations without using negative samples. It combines intra-view variance and covariance regularization with inter-view invariance and cross-correlation alignment. The objective integrates principles used in VICReg \cite{bardes2022vicreg} and Barlow Twins \cite{zbontar2021barlow}, within a unified formulation that jointly optimizes intra-view and inter-view criteria. We weight the different regularization terms using non-negative scalars $\lambda_{\mathrm{intra}}$, $\lambda_{\mathrm{inv}}$, and $\lambda_{\mathrm{corr}}$, which control their relative contribution to the total loss.

\paragraph{Intra-view regularization.}
We apply variance and covariance regularization independently to each view. Let $Z^{(j)}\in\mathbb{R}^{N}$ denote the $j$-th embedding dimension across the batch, and let $\mathrm{Std}(Z^{(j)})$ denote its empirical standard deviation. The variance term enforces a minimum standard deviation $\gamma>0$:
\begin{equation}
\mathcal{L}_{\mathrm{var}}(Z)
=
\frac{1}{D}\sum_{j=1}^{D}
\max\!\Bigl(0,\ \gamma - \mathrm{Std}(Z^{(j)})\Bigr),
\label{eq:sicova_var}
\end{equation}
where the standard deviation is computed with a small constant added inside the square root for numerical stability.

The covariance regularization penalizes correlations between different embedding dimensions. Let $C_Z\in\mathbb{R}^{D\times D}$ denote the sample covariance matrix of $Z$ (computed after centering each dimension). The covariance loss is:
\begin{equation}
\mathcal{L}_{\mathrm{cov}}(Z)
=
\frac{1}{D}\sum_{i\neq j}(C_Z)_{ij}^{2}.
\label{eq:sicova_cov}
\end{equation}

We apply these terms to both views and sum:
\begin{equation}
\mathcal{L}_{\mathrm{intra}}(Z,Z')
=
\mathcal{L}_{\mathrm{var}}(Z)+\mathcal{L}_{\mathrm{var}}(Z')
+
\mathcal{L}_{\mathrm{cov}}(Z)+\mathcal{L}_{\mathrm{cov}}(Z').
\label{eq:sicova_intra}
\end{equation}

\paragraph{Inter-view regularization.}
To enforce invariance across views, we minimize the mean squared distance between paired embeddings:
\begin{equation}
\mathcal{L}_{\mathrm{inv}}(Z,Z')
=
\frac{1}{N}\sum_{i=1}^{N}\| z_i - z'_i \|_2^2.
\label{eq:sicova_inv}
\end{equation}

To promote alignment of corresponding embedding dimensions while discouraging redundancy across dimensions, we use a cross-correlation criterion. Let $\tilde Z$ and $\tilde Z'$ denote column-wise centered and $l_2$-normalized embeddings, and define the cross-correlation matrix:
\begin{equation}
\mathcal{R}
=
\frac{1}{N}\tilde Z^{\top}\tilde Z'
\in \mathbb{R}^{D\times D}.
\label{eq:sicova_corrmat}
\end{equation}
The correlation loss is:
\begin{equation}
\mathcal{L}_{\mathrm{corr}}(Z,Z')
=
\sum_{i}\bigl(1-\mathcal{R}_{ii}\bigr)^2
+
\sum_{i\neq j}\mathcal{R}_{ij}^{2}.
\label{eq:sicova_corr}
\end{equation}

\paragraph{Overall objective.}
The SiCoVa objective is a weighted sum of intra-view and inter-view terms:
\begin{equation}
\begin{aligned}
\mathcal{L}_{\mathrm{SiCoVa}}
&=
\lambda_{\mathrm{intra}}\mathcal{L}_{\mathrm{intra}}(Z,Z')
+
\lambda_{\mathrm{inv}}\mathcal{L}_{\mathrm{inv}}(Z,Z') \\
&\quad
+
\lambda_{\mathrm{corr}}\mathcal{L}_{\mathrm{corr}}(Z,Z').
\end{aligned}
\label{eq:sicova_total}
\end{equation}

All weights are fixed across experiments.

\subsection{Contrastive Baseline: Triplet Loss}

As a contrastive baseline, we use a batch-all triplet loss \cite{hermans2017triplet}. For each anchor embedding $z_i$ from $Z$, the positive sample is the corresponding embedding $z'_i$ from $Z'$, and all other embeddings $z'_k$ with $k \neq i$ act as negatives. With margin $m>0$ and hinge function $[\cdot]_+=\max(\cdot,0)$, the loss is:
\begin{equation}
\mathcal{L}_{\mathrm{triplet}}
=
\frac{1}{N(N-1)}\sum_{i=1}^{N}\sum_{k\neq i}
\Bigl[
\lVert z_i - z'_i \rVert_2
-
\lVert z_i - z'_k \rVert_2
+
m
\Bigr]_+.
\label{eq:triplet}
\end{equation}
This objective emphasizes inter-instance discrimination and serves as a contrastive reference to SiCoVa.

\subsection{Downstream Fine-Tuning}

Each SSL checkpoint is fine-tuned on labeled diabetic retinopathy data using an identical supervised protocol. We use fixed training, validation, and calibration splits across all experiments to ensure fair comparison. The downstream task is multi-class DR severity classification.

Class activation map (CAM) based refinement is applied during downstream fine-tuning to improve localization and raw classification performance. However, CAM refinement is \emph{not} used when evaluating selective prediction and abstention. This separation ensures that abstention behavior and confidence-based deferral are driven by the learned representations and calibrated probabilities, rather than by additional spatial regularization introduced during fine-tuning.

\subsection{Calibration and Selective Prediction}

After fine-tuning, predicted class probabilities are calibrated using temperature scaling on a held-out calibration split \cite{guo2017calibration}. Calibration is performed independently for each SSL checkpoint using identical data and procedures. In particular, we fit a single temperature parameter on the calibration split and apply it to logits at inference time to obtain calibrated class probabilities.

Selective prediction is implemented via confidence-threshold abstention \cite{geifman2017selective} as a post hoc decision rule, and does not affect training. For an input $x$, let $p_{\max}(x)=\max_k p(y=k\mid x)$ denote the maximum calibrated predicted probability. Given a threshold $\tau$, the model predicts if $p_{\max}(x)\ge \tau$ and abstains otherwise. Coverage is defined as the fraction of samples that are retained (not abstained on).

To support comparisons across checkpoints and methods, we report selective results at a fixed coverage of 70\%. This operating point was chosen as a clinically realistic trade-off that retains the majority of predictions while still allowing substantial abstention of uncertain cases for expert review. Concretely, for each checkpoint, we sweep a fixed grid of thresholds $\tau$ and select the operating point whose coverage is closest to 70\%, then report selective accuracy and selective macro-F1 at that operating point. We additionally observe qualitatively similar trends across neighboring operating points in the risk-coverage analysis.

\section{Experimental Setup}

This section describes the controlled experimental setup used to isolate the effect of SSL pretraining length on confidence behavior and selective prediction.
\subsection{Datasets}

We study diabetic retinopathy (DR) severity classification using retinal fundus images. Our primary evaluation benchmark is APTOS-19, a widely used dataset in the DR literature that provides relatively consistent annotations compared to other public benchmarks \cite{tsiknakis2021reviewdr}. APTOS-19 contains five disease severity levels, ranging from no DR to proliferative DR.

For representation learning, models are pretrained on EyePACS using unlabeled images only. EyePACS provides large-scale and diverse retinal data and is commonly used for learning general-purpose fundus representations. All downstream fine-tuning and primary evaluation are performed on APTOS-19.

To assess robustness and generalization, we additionally evaluate performance on Messidor and a 7-class fundus dataset. These datasets are not used during self-supervised pretraining. For downstream evaluation, we fine-tune and evaluate models on each dataset using the same controlled protocol and labeled-data regimes as for APTOS-19.

Table~\ref{tab:datasets} summarizes the datasets used in our experiments.

\begin{table}[t]
\centering

\small
\setlength{\tabcolsep}{3pt}
\renewcommand{\arraystretch}{1.15}
\begin{tabular}{>{\raggedright\arraybackslash}p{2.65cm}ccc}
\toprule
Dataset & Role & Classes & Images \\
\midrule
\parbox[t][2.6\baselineskip][t]{\linewidth}{EyePACS\\\cite{Kaggle2015EyePACS}}
& SSL pretraining & -- & 35{,}000 \\

\parbox[t][2.6\baselineskip][t]{\linewidth}{APTOS-19\\\cite{aptos2019kaggle}}
& Fine-tuning & 5 & 3{,}662 \\

\parbox[t][2.6\baselineskip][t]{\linewidth}{Messidor\\\cite{Messidor2}}
& Fine-tuning & 4 & 1{,}200 \\

\parbox[t][2.6\baselineskip][t]{\linewidth}{7-class Fundus\\\cite{castillo_benitez_2021_fundus_dr}}
& Fine-tuning & 7 & 757 \\
\bottomrule
\end{tabular}
\caption{Datasets used for pretraining, fine-tuning, and evaluation.}
\label{tab:datasets}

\end{table}

\subsection{Data Splits and Training Protocol}

For downstream training, we use fixed training, validation, and calibration splits for all SSL checkpoints. The calibration split is used exclusively for temperature scaling, while the validation split is used for evaluation both with and without abstention. Identical splits are used across all experiments to ensure fair comparison across pretraining lengths and methods.

We evaluate multiple labeled data regimes by varying the fraction of APTOS-19 training data used for fine-tuning. This allows us to assess whether the impact of SSL pretraining length on reliability persists under limited supervision, which is common in medical imaging settings.

\subsection{Evaluation Metrics}

We report standard downstream performance without abstention using accuracy and macro-F1. To assess reliability under selective prediction, we additionally report selective accuracy and selective macro-F1 after applying confidence-threshold abstention. Coverage is defined as the fraction of samples for which the model produces a prediction (i.e., not abstained on). For fair comparison across checkpoints and methods, we report selective results at a fixed coverage of $70\%$.

Macro-averaged metrics are essential in diabetic retinopathy screening due to class imbalance and severity-dependent rejection effects. In particular, selective prediction changes the evaluated set and can skew class frequencies, making accuracy alone potentially misleading. We therefore use selective macro-F1 as the primary metric for ranking checkpoints at the target coverage.

Beyond aggregate scores, we analyze coverage--risk trends and class-wise rejection patterns to characterize how pretraining length reshapes deployment-relevant trade-offs and abstention behavior across disease stages.

\subsection{Implementation Details}

All experiments use the same architecture and optimization settings across SSL checkpoints. Training and evaluation are performed on a single NVIDIA A100 GPU. Self-supervised pretraining requires approximately 10 hours per model, while downstream fine-tuning takes approximately 1.25 hours per model. At inference time, the abstention mechanism introduces negligible computational overhead, as it relies solely on confidence-based post-processing.

\section{Results}

We now present experimental results analyzing the effect of self-supervised pretraining length on downstream performance and reliability in diabetic retinopathy screening. We first examine standard classification performance without abstention, and then analyze selective prediction behavior under a fixed operating point.

\subsection{Downstream Performance }

\begin{figure}[!htbp]
\centering
\includegraphics[width=\columnwidth]{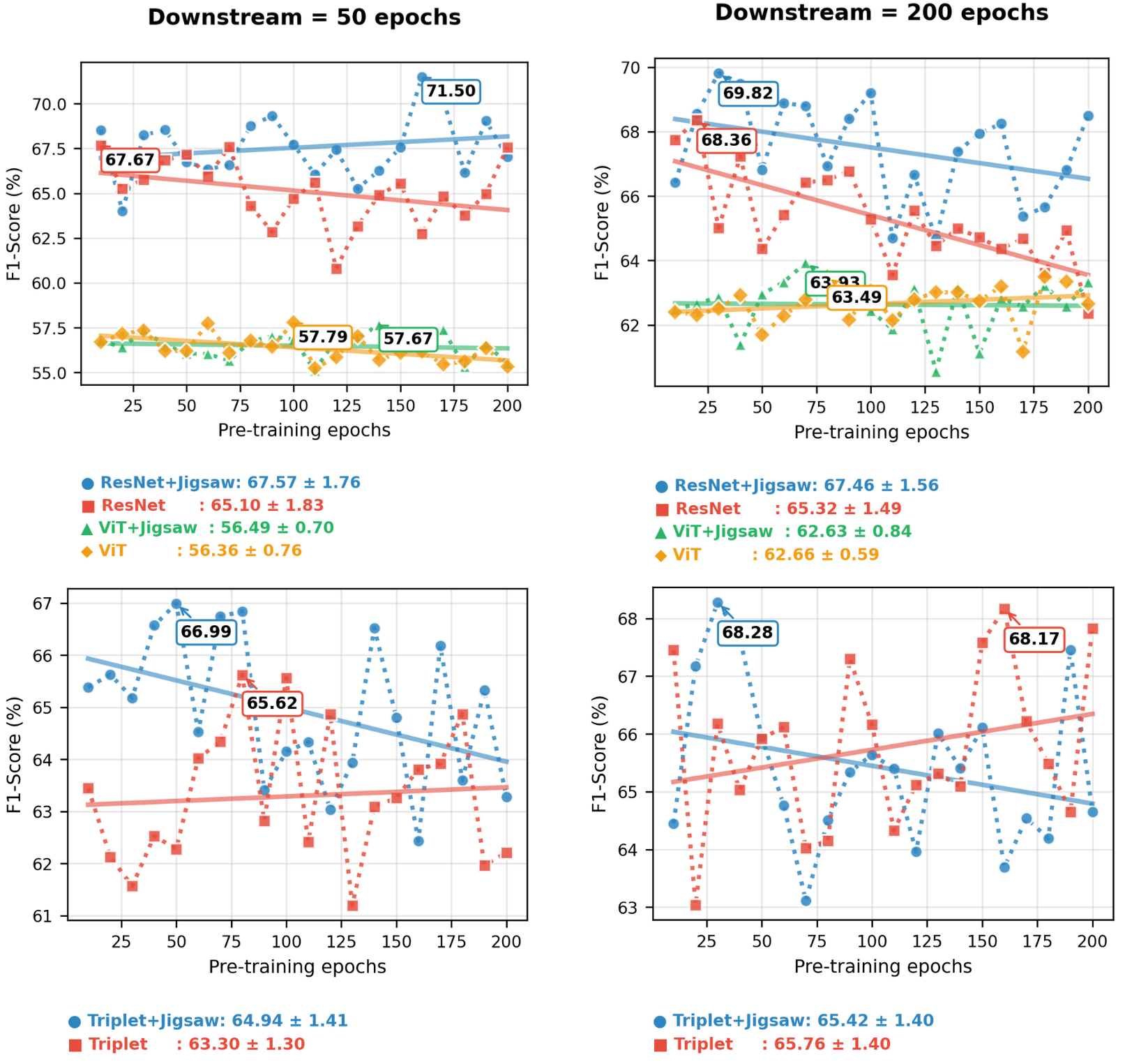}
\caption{Downstream macro-F1 across SSL pretraining checkpoints for SiCoVa and Triplet, shown for two fine-tuning durations (epoch 50 and epoch 200).}
\label{fig:f1_vs_pretrain_4panel}
\end{figure}

We begin by analyzing downstream classification performance as a function of SSL pretraining length. Figure~\ref{fig:f1_vs_pretrain_4panel} shows downstream macro-F1 across SSL checkpoints for SiCoVa and Triplet, evaluated at downstream fine-tuning epochs 50 and 200 on APTOS-19 dataset.

Across methods, macro-F1 improves rapidly during early stages of pretraining and then saturates. Extending pretraining beyond this point yields marginal or inconsistent gains, and in some cases mild degradation. This trend is consistent across both fine-tuning durations, indicating that longer downstream training does not compensate for saturation in representation learning.

These observations align with prior work showing that SSL representations converge early with respect to downstream accuracy \cite{truong2021transferability}. While some checkpoint-level variability is expected due to stochastic fine-tuning, we focus next on whether pretraining length systematically affects reliability.

Table~\ref{tab:sicova_all} reports downstream performance of SiCoVa across datasets and labeled data regimes. While accuracy and macro-F1 improve with increased supervision, gains diminish at higher data fractions, particularly on APTOS.

\begin{table}[h]
\centering

\small
\setlength{\tabcolsep}{5pt}
\renewcommand{\arraystretch}{1.05}

\begin{tabular}{llrcr}
\toprule
Dataset & Regime & Acc (\%) & Macro-F1 (\%) & QWK \\
\midrule
APTOS 
& 10\%   & 79.0 & 53.0 & 0.86 \\
& 50\%   & 81.0 & 64.0 & 0.88 \\
& 100\%  & \textbf{85.27} & \textbf{71.5} & \textbf{0.91} \\
\midrule
Messidor 
& 10\%   & 54.0 & 54.0 & 0.24 \\
& 50\%   & 69.0 & 69.0 & 0.77 \\
& 100\%  & \textbf{71.0} & \textbf{71.0} & \textbf{0.77} \\
\midrule
7-class Fundus 
& 10\%   & 67.0 & 67.0 & 0.82 \\
& 50\%   & 86.0 & 86.0 & 0.95 \\
& 100\%  & \textbf{88.0} & \textbf{88.0} & \textbf{0.97} \\
\bottomrule
\end{tabular}
\caption{SiCoVa results across datasets and labeled data regimes. Metrics are computed without abstention.}
\label{tab:sicova_all}
\end{table}

To contextualize these results, Table~\ref{tab:aptos_all_noabs} compares SiCoVa against Triplet, supervised baselines, and prior approaches on APTOS without abstention. SiCoVa achieves the strongest overall performance, particularly when combined with CAM refinement. We next examine whether these accuracy differences translate to improved reliability under selective prediction.

\begin{table}[t]
\centering
\small
\setlength{\tabcolsep}{3.5pt}
\renewcommand{\arraystretch}{1.05}

\begin{tabular}{llrrrrr}
\toprule
Method & Backbone & Acc & F1 & Prec. & Rec. & QWK \\
\midrule
\multirow{5}{*}{SiCoVa}
& R50 + J + C   & \textbf{85.27} & \textbf{71.50} & \textbf{75.29} & \textbf{69.12} & \textbf{0.91} \\
& R50 + J        & 79.00 & 57.00 & 59.00 & 56.00 & 0.85 \\
& R50             & 83.77 & 67.56 & 71.60 & 65.17 & 0.90 \\
& ViT-B/16 + J   & 79.67 & 57.39 & 68.72 & 54.50 & 0.82 \\
& ViT-B/16        & 79.95 & 57.79 & 69.02 & 54.92 & 0.83 \\
\midrule
Triplet
& R50 + J + C   & 83.90 & 66.74 & 69.55 & 65.35 & 0.84 \\
& R50             & 84.17 & 62.53 & 72.26 & 62.94 & 0.87 \\
\midrule
Supervised
& R50             & 80.27 & 61.77 & 64.89 & 60.07 & 0.85 \\
& ViT-B/16        & 83.00 & 65.00 & 72.00 & 62.00 & 0.89 \\
\midrule
ClementP$^{1}$
& R50             & 51.00 & 28.00 & 35.00 & 38.00 & 0.66 \\
& ViT-B/16        & 69.00 & 51.00 & 54.00 & 54.00 & 0.88 \\
\bottomrule
\end{tabular}
\caption{APTOS classification performance across SSL and supervised baselines. (R50 = ResNet50, ViT-B/16 = ViT-base (16×16), J = Jigsaw SSL, C = CAM refinement.)}
\label{tab:aptos_all_noabs}

\end{table}

\footnotetext[1]{ClementP fundus grading collection: \url{https://huggingface.co/collections/ClementP/fundus-grading}}

\subsection{Selective Prediction at Fixed Coverage}

We next evaluate selective prediction using confidence-based abstention. All selective results are reported at a fixed coverage of $70\%$, reflecting a realistic screening setting in which uncertain cases are deferred for expert review.

Table~\ref{tab:selective_results} summarizes selective performance at this operating point for representative SSL checkpoints (with pretraining epochs shown). While selective accuracy remains high across checkpoints, selective macro-F1 varies substantially, even among models with similar downstream performance. Notably, longer SSL pretraining does not consistently improve selective macro-F1: intermediate checkpoints can outperform later checkpoints despite comparable accuracy, suggesting that extended pretraining reshapes confidence allocation rather than raw predictive performance.

\begin{table}[h]
\centering
\small
\setlength{\tabcolsep}{5pt}
\renewcommand{\arraystretch}{1.05}
\begin{tabular}{l l c c r}
\toprule
Method & Checkpoint tier & Sel. Acc. (\%) & Sel. F1 (\%) & QWK \\
\midrule
\multirow{3}{*}{SiCoVa}
& Best (ep-110)   & \textbf{94.69} & \textbf{78.65} & \textbf{0.95} \\
& Median (ep-150) & 93.93 & 75.57 & 0.94 \\
& Weak (ep-60)    & 92.68 & 68.74 & 0.94 \\
\midrule
\multirow{3}{*}{Triplet}
& Best (ep-170)   & 92.31 & 46.56 & 0.92 \\
& Median (ep-40)  & 93.16 & 37.45 & 0.93 \\
& Weak (ep-20)    & 91.52 & 36.93 & 0.90 \\
\bottomrule
\end{tabular}
\caption{Selective prediction performance at $70\%$ coverage. }
\label{tab:selective_results}

\end{table}

\begin{figure}[h]
\centering
\includegraphics[width=\columnwidth]{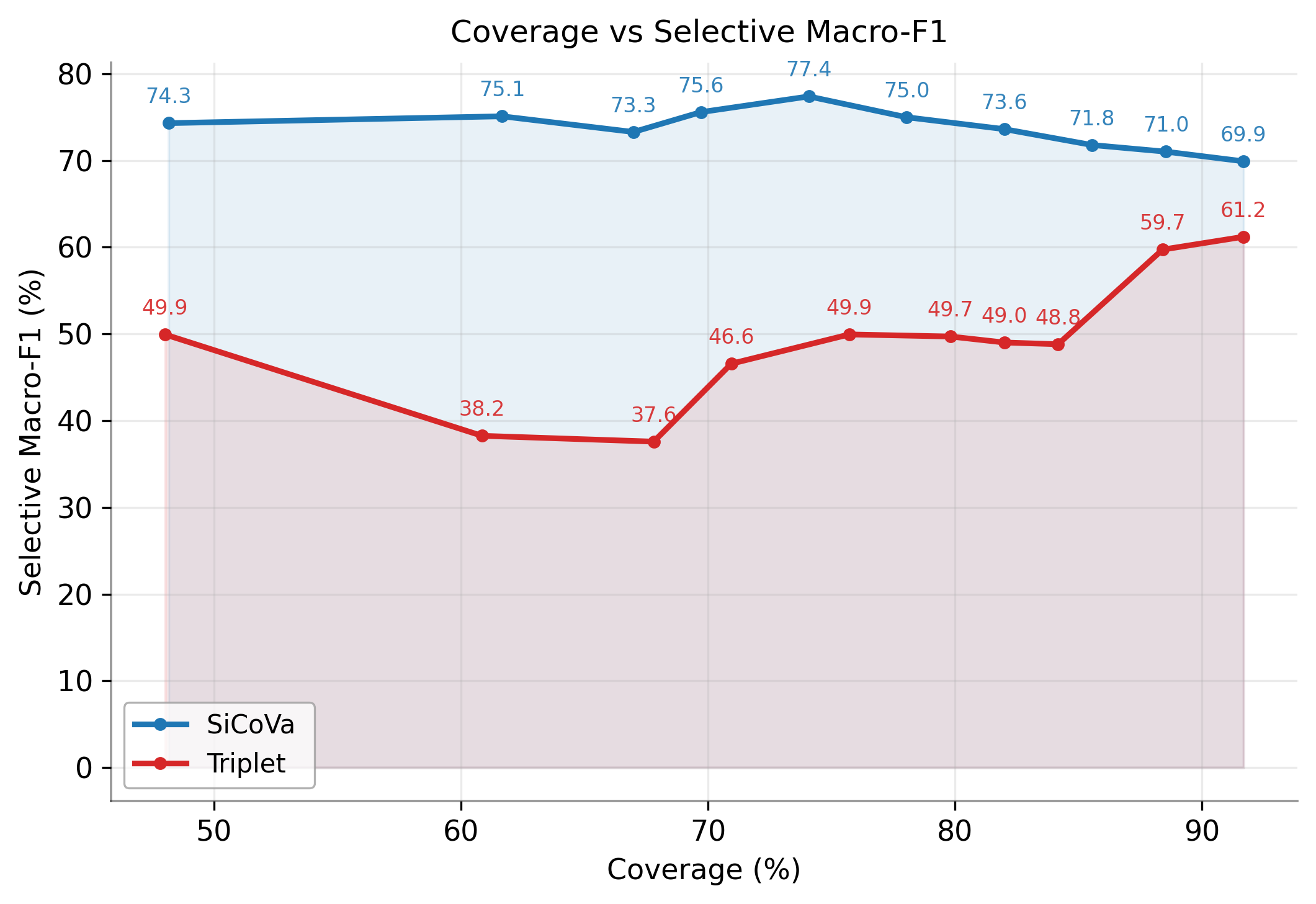}
\caption{Selective prediction comparison between SiCoVa and Triplet under confidence-based abstention. We sweep the confidence threshold to obtain a risk-coverage curve, and highlight the operating point corresponding to $70\%$ coverage used in Table~\ref{tab:selective_results}.}
\label{fig:coverage_vs_selective_f1}
\end{figure}

Figure~\ref{fig:coverage_vs_selective_f1} complements Table~\ref{tab:selective_results} by sweeping the confidence threshold to vary coverage and obtain a risk--coverage curve. SiCoVa consistently achieves higher selective macro-F1 across operating points, whereas Triplet is more sensitive to the choice of coverage. The highlighted point corresponds to the $70\%$ coverage operating point used in the table.

\subsection{Accuracy versus Macro-Averaged Metrics under Abstention}

As confidence thresholds increase, selective accuracy rises sharply due to the removal of ambiguous samples and minority-class instances. In contrast, selective macro-F1 improves more gradually and often plateaus.

This discrepancy arises because accuracy is dominated by majority classes and benefits from aggressive rejection, whereas macro-F1 penalizes uneven class-wise performance. As a result, selective macro-F1 provides a more informative measure of reliability under abstention for imbalanced screening tasks.

\subsection{Representation Structure and Qualitative Analysis}

\begin{figure}[h]
\centering
\includegraphics[width=\columnwidth]{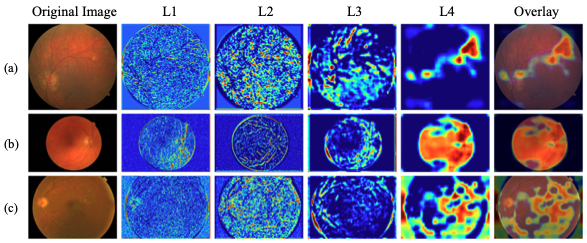}
\caption{Grad-CAM visualizations for SiCoVa across three downstream datasets, contrasting accepted and rejected cases [Dataset: (a) 7-Class Fundus, (b) Messidor, and (c) Aptos-19].}
\label{fig:gradcam_three_datasets}
\end{figure}

Figure~\ref{fig:gradcam_three_datasets} provides qualitative Grad-CAM visualizations across APTOS, Messidor, and the 7-class fundus dataset. Accepted predictions typically exhibit more coherent activation patterns, while rejected cases more often show diffuse or unstable responses. These visualizations serve as qualitative sanity checks rather than localization claims.

\begin{figure*}[t]
\centering
\includegraphics[width=0.85\textwidth]{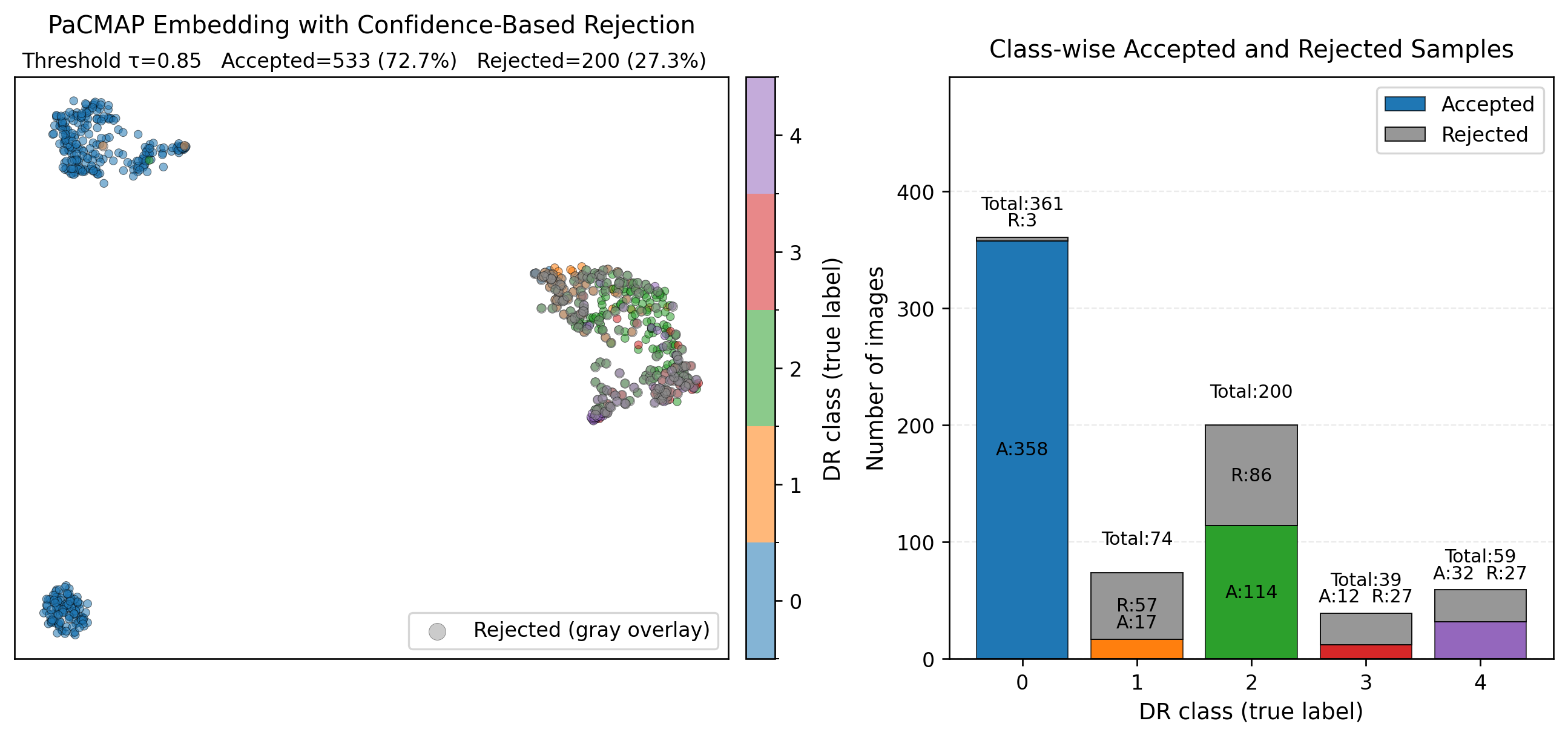}
\caption{Representation structure and severity-dependent abstention using PaCMAP and class-wise acceptance statistics at 70\% coverage.}
\label{fig:pacmap_and_classwise_reject}
\end{figure*}

Figure~\ref{fig:pacmap_and_classwise_reject} visualizes representation structure and severity-dependent rejection using PaCMAP and class-wise acceptance statistics. Disease severity classes form a continuous and partially overlapping manifold, reflecting the progressive nature of diabetic retinopathy. After abstention, accepted samples occupy denser and more class-consistent regions, while rejected samples concentrate near inter-class boundaries rather than forming isolated outliers, suggesting that abstention primarily filters semantically ambiguous cases.

\subsection{Cross-Dataset Transfer}

Finally, we evaluate selective prediction on Messidor and the 7-class fundus dataset. While downstream accuracy remains relatively stable across SSL checkpoints, selective macro-F1 continues to vary, indicating that the impact of pretraining length on confidence behavior generalizes beyond a single benchmark.

Overall, these results demonstrate that SSL pretraining length has a limited effect on downstream accuracy but a pronounced impact on confidence calibration and selective prediction behavior, with direct implications for safe deployment in clinical screening systems.

\section{Discussion}

Our results demonstrate that the length of self-supervised pretraining plays a meaningful role in shaping confidence behavior and selective prediction performance in diabetic retinopathy screening. While downstream accuracy often saturates early, reliability-aware evaluation reveals substantial variation across SSL checkpoints that is not captured by standard metrics. In this section, we discuss the implications of these findings for representation learning, clinical interpretation, and deployment.

\subsection{Why Longer Pretraining Does Not Guarantee Better Reliability}

A common assumption in SSL-based medical imaging is that longer pretraining is either beneficial or, at worst, neutral once downstream accuracy has converged. Our results challenge this assumption. Although extended pretraining does not consistently degrade standard downstream performance, it can still reshape confidence distributions, which may affect selective macro-F1 at a fixed coverage. We examine this relationship directly by reporting selective performance across pretraining epochs in the subsequent analysis.

One plausible explanation is that later stages of SSL pretraining emphasize invariances that improve global discrimination while compressing features relevant for borderline or minority classes. In diabetic retinopathy, disease severity evolves along a continuum rather than forming well-separated categories. As a result, representations that appear strong under accuracy-based evaluation may still yield poorly calibrated confidence near class boundaries. These effects remain largely invisible unless selective prediction behavior is explicitly analyzed.

Importantly, our findings do not suggest that longer pretraining is inherently harmful. Rather, they indicate that pretraining length interacts with downstream reliability in non-monotonic ways, and that accuracy alone is insufficient to identify favorable operating points.

\subsection{Clinical Interpretation of Accepted and Rejected Cases}

To assess the clinical plausibility of abstention decisions, a clinician inspected a subset of accepted and rejected predictions. Rejected cases often exhibited factors known to complicate automated grading, including decentration, shadowing, media opacities, and confounding retinal conditions, consistent with criteria for non-gradable images in screening workflows. The clinician also noted occasional inconsistencies within the accepted set and a small number of rejected images of acceptable quality. This underscores that confidence-based abstention reflects model uncertainty, not clinical ground truth or image quality, and may conservatively reject borderline but usable cases. Overall, abstention should be interpreted as decision support rather than a replacement for expert judgment \cite{defauw2018clinically}.

\subsection{Accuracy versus Reliability in Screening Systems}

Our analysis highlights limitations of accuracy as a deployment-oriented metric. Under selective prediction, accuracy can increase as uncertain samples are rejected even when performance on clinically relevant minority or borderline classes degrades. Macro-averaged metrics expose this effect by penalizing uneven class-wise performance and severity-dependent rejection. Selective macro-F1 therefore, provides a more balanced assessment of reliability under abstention and aligns better with equitable performance across severity levels.

\subsection{Implications for Deployment and Evaluation}

First, SSL pretraining length should be treated as a design choice with direct consequences for confidence behavior. While selective performance varies across checkpoints, we do not observe a clear monotonic trend with pretraining length, motivating further experiments. Second, benchmarking SSL methods solely on downstream accuracy can miss clinically relevant differences in uncertainty handling. Third, confidence-based abstention is a simple deferral mechanism, but its behavior should be assessed across operating points and disease severities.

\section{Conclusion}

We study how self-supervised pretraining length influences reliability and selective prediction in diabetic retinopathy screening. Although downstream accuracy saturates early, confidence behavior and selective macro-F1 vary substantially across SSL checkpoints, affecting deployment decisions under abstention. These findings motivate reliability-aware evaluation and treating pretraining length as a primary design choice.

\paragraph{Limitations.}
We focus on a single SSL framework and a limited set of architectures, and the clinical inspection is qualitative and based on a small sample. We rely on confidence-based abstention rather than learned deferral. In addition, we do not perform statistical significance testing across checkpoints or repeated fine-tuning runs. Future work should extend to additional SSL objectives and architectures, analyze reliability trends across multiple random seeds, and include larger-scale human-in-the-loop evaluation.

\section*{Acknowledgments}

This research has been funded by the Federal Ministry of Education and
Research of Germany and the state of North-Rhine Westphalia as part of the
Lamarr Institute for Machine Learning and Artificial Intelligence.

\bibliographystyle{named}
\bibliography{ijcai26}

\end{document}